\documentclass[10pt,twocolumn,letterpaper]{article}

\usepackage[pagenumbers]{cvpr} 

\definecolor{cvprblue}{rgb}{0.21,0.49,0.74}
\usepackage[pagebackref,breaklinks,colorlinks,allcolors=cvprblue]{hyperref}

\usepackage{multirow}
\usepackage{arydshln}

\def\paperID{16281} 
\def\confName{CVPR}
\def\confYear{2025}

\title{MVUDA: Unsupervised Domain Adaptation for Multi-view Pedestrian Detection}

\author{Erik Brorsson$^{\star \dagger}$ \qquad Lennart Svensson$^{\dagger}$ \qquad Kristofer Bengtsson$^{\star}$ \qquad Knut Åkesson$^{\dagger}$ \\
$^{\star}$Global Trucks Operations, Volvo Group, Gothenburg, Sweden \\
$^{\dagger}$Department of Electrical Engineering, Chalmers University of Technology, Gothenburg, Sweden \\
{\tt\small \{erik.brorsson, kristofer.bengtsson\}@volvo.com, \{lennart.svensson, knut.akesson\}@chalmers.se} 
}

\begin{document}
\maketitle
\begin{abstract}

We address multi-view pedestrian detection in a setting where labeled data is collected using a multi-camera setup different from the one used for testing. While recent multi-view pedestrian detectors perform well on the camera rig used for training, their performance declines when applied to a different setup.
To facilitate seamless deployment across varied camera rigs, we propose an unsupervised domain adaptation (UDA) method that adapts the model to new rigs without requiring additional labeled data. Specifically, we leverage the mean teacher self-training framework with a novel pseudo-labeling technique tailored to multi-view pedestrian detection. This method achieves state-of-the-art performance on multiple benchmarks, including MultiviewX$\rightarrow$Wildtrack. 
Unlike previous methods, our approach eliminates the need for external labeled monocular datasets, thereby reducing reliance on labeled data. Extensive evaluations demonstrate the effectiveness of our method and validate key design choices. By enabling robust adaptation across camera setups, our work enhances the practicality of multi-view pedestrian detectors and establishes a strong UDA baseline for future research.

\end{abstract}
    
\section{Introduction}
\label{sec:intro}

    Multi-view detection aims to detect objects from a set of images captured simultaneously by multiple cameras, each providing a distinct view of the same scene. Using multiple views allows for greater robustness to occlusions and facilitates inferring 3D properties of objects, which can be challenging with a single camera. In this paper, we focus on multi-view pedestrian detection, where the goal is to generate an occupancy map in bird's-eye-view (BEV) from images captured by multiple stationary cameras. This task is relevant in applications like surveillance \cite{ferryman2009pets2009}, robotics \cite{coates2010multi}, sports analytics \cite{ren2010multi}, and autonomous mobile robot control~\cite{10185133}.

    \begin{figure}[h!]
        \centering
        \includegraphics[width=1.0\linewidth]{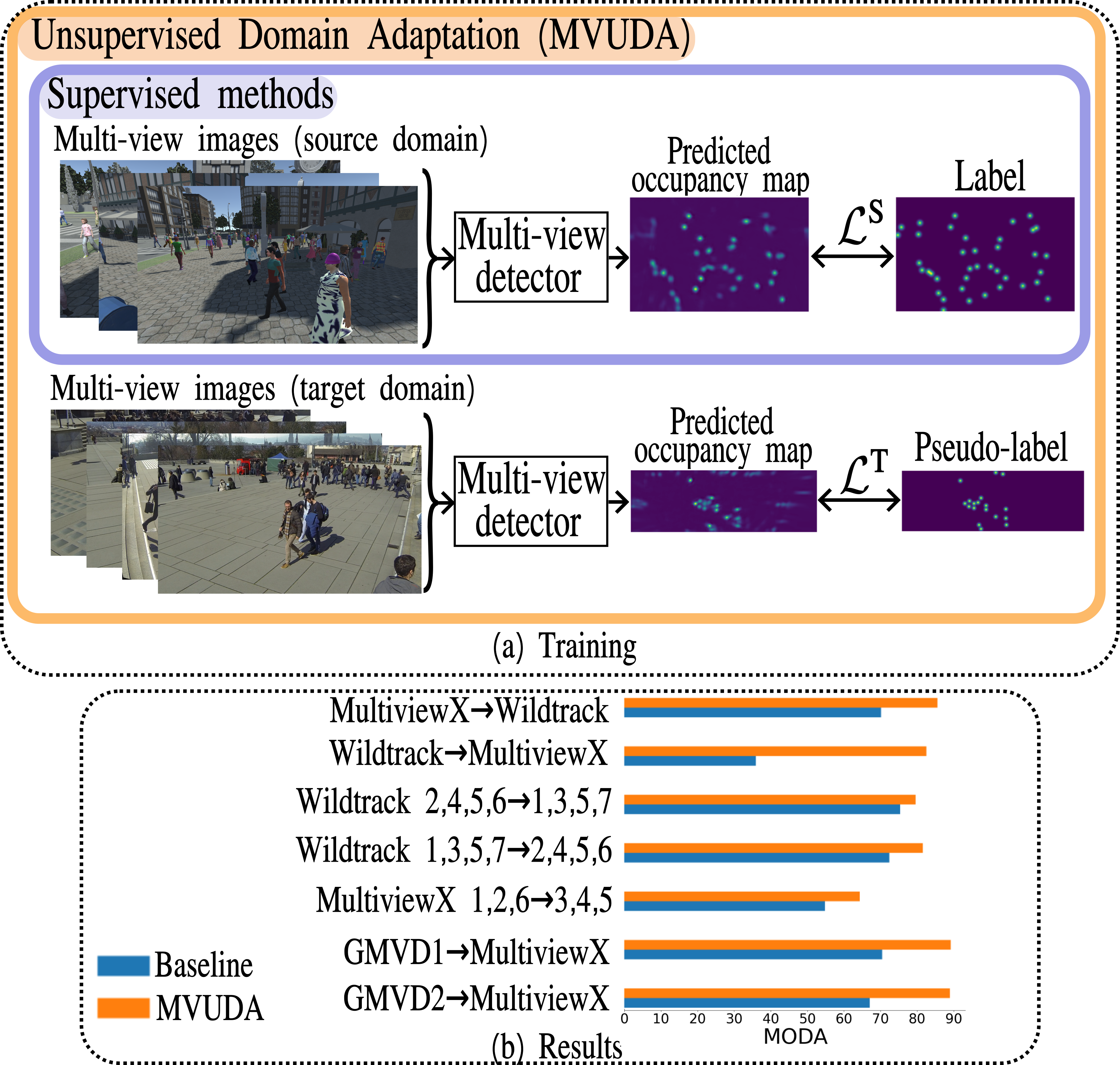}
        \caption{
        Since labeled multi-view datasets are scarce, current methods for multi-view pedestrian detection that rely on labeled (source) datasets for training do not perform well on new camera setups (target). We consider unsupervised domain adaptation (a) where labeled source data alongside pseudo-labeled target data is used for training, greatly improving the model's performance on multiple benchmarks (b).
        }
        \label{fig:enter-label}
    \end{figure}

    Recent methods for multi-view pedestrian detection consider all input images jointly to learn a dense BEV feature map \cite{mvdet, gmvd, shot, 3drom, mvaug, mvfp, hou2021multiview}. This BEV representation is then refined, typically with convolutional layers, to obtain a probabilistic occupancy map (POM), from which detections can be extracted. 
    Although these methods have achieved impressive results, they rely on labeled multi-view datasets, which are typically scarce due to the costs of multi-camera setups and image annotation.
    In practice, labeled data is typically limited to simulations or a single real-world camera rig, leading to overfitting and poor generalization across different camera setups.

    Collecting unlabeled data from the real-world test setup, however, is relatively straight-forward, making unsupervised domain adaptation (UDA) a promising solution to the generalization challenges in multi-view detection. UDA is well established for monocular perception tasks such as image classification, semantic segmentation, and object detection, with mean teacher self-training as a popular approach \cite{deng2021unbiased,hoyer2022daformer,li2022cross}. This approach trains a student model on unlabeled data using pseudo-labels generated by a mean teacher \cite{tarvainen2017mean}, an exponential moving average of the student’s parameters. However, to the best of our knowledge, Lima \etal \cite{lima2023toward, lima2024mean} constitute the only works to explore UDA in multi-view pedestrian detection. In their approach, they adapt a multi-view detector through self-training, but rely on a pre-trained external detector based on large, labeled monocular datasets, limiting practicality for applications without access to such resources.   

    We address this gap by considering a strict UDA setting that excludes any external labeled dataset or pre-trained detector. Apart from its practical relevancy due to restrictive licensing of datasets and derived detectors, it is also conceptually interesting as it opens possibilities to extend the framework to new object types in the future. We build on mean teacher self-training, adapting it for multi-view pedestrian detection and identifying key success factors for the strict UDA settings.  
    Importantly, we propose a novel post-processing method to enhance pseudo-label reliability, significantly improving self-training efficacy.
    Our method achieves state-of-the-art performance across multiple benchmarks. Furthermore, while recent works primarily focus on bridging simulated and real-world domains, few consider the challenges posed by changing camera configurations. To facilitate this, we introduce two new benchmarks specifically for cross-camera rig adaptation.   
     
    Our contributions can be summarized as follows:
    \begin{enumerate}
        \item We unveil the potential of self-training for multi-view pedestrian detection under a strict UDA setting and develop a state-of-the-art method for this problem.
        \item We propose a simple yet effective post-processing method that improves pseudo-label reliability and thereby the effectiveness of self-training.
        \item We demonstrate the efficacy of our method on multiple established benchmarks and on two new benchmarks, which we introduce to specifically address cross-camera rig adaptation.
    \end{enumerate}

\section{Related Work}

    \subsection{Multi-view pedestrian detection}
    Multi-view pedestrian detection aims to utilize cameras with different viewpoints to enable more robust detection and localization in 3D than what is possible with a single camera. Early methods relied on background subtraction in each view and inferred 3D ground plane positions using graphical models combined with Bayesian inference \cite{fleuret2007pom, alahi2011sparsity, peng2015robust}. 
    Since background subtraction is not sufficiently discriminative in crowded scenes, many later works replaced this component with more advanced methods of monocular perception, such as 2D bounding box detection \cite{lima20223d, lopez2022semantic, roig2011conditional}, human pose estimation \cite{lima20223d}, or instance segmentation \cite{qiu2024ppm}. These methods also proposed alternative ways to fuse individual detections, such as projecting detections onto a ground plane and grouping them based on Euclidean proximity \cite{lima20223d, lopez2022semantic, qiu2024ppm}, or employing Conditional Random Fields (CRF) \cite{roig2011conditional}. However, because these methods rely on monocular perception, any deficiencies in the individual views can degrade overall performance.  
    
    
    In contrast, end-to-end methods consider all input images jointly, enabling a more comprehensive understanding of correspondences across views. Early methods processed each view with a Convolutional Neural Network (CNN) to extract features and then applied either a Multilayer Perceptron (MLP) \cite{chavdarova2017deep} or CRF \cite{baque2017deep} to generate detections by jointly considering these features. Recently, MVDet \cite{mvdet} introduced a new approach by projecting features from individual views into a bird's-eye view (BEV) through a perspective transformation, creating dense feature maps in BEV. Many recent methods build on this idea through improved perspective view feature extraction \cite{lee2023multi}, enhanced feature aggregation in BEV \cite{gmvd, shot, mvfp}, modified decoders \cite{teepe2024earlybird, hou2021multiview}, and multi-view-specific data augmentation techniques \cite{mvaug, 3drom}.
    While these approaches continue to push the state-of-the-art in multi-view pedestrian detection, they require labeled multi-view datasets for training and typically fail to generalize well to new camera setups. In this work, we aim to relax the dependency on labeled multi-view data, making these methods more useful in practice.

    \subsection{Unsupervised Domain Adaptation (UDA)}
    Given a labeled dataset from a source domain and an unlabeled dataset from a target domain, Unsupervised Domain Adaptation (UDA) aims to transfer knowledge from the source to the target, allowing models to generalize to new data distributions without additional labels. UDA has been widely applied in computer vision tasks, including image classification \cite{ganin2016domain, long2015learning, saito2018maximum}, semantic segmentation \cite{hoffman2016fcns,hoffman2018cycada,gong2019dlow,tsai2018learning}, and object detection \cite{li2022cross,deng2021unbiased,cai2019exploring,cao2023contrastive}. Recent UDA methods largely follow two approaches: adversarial learning and self-training. 
    Adversarial learning seeks to create domain-invariant input \cite{hoffman2018cycada,gong2019dlow,deng2021unbiased}, output \cite{saito2018maximum,tsai2018learning} or features \cite{ganin2016domain,hoffman2016fcns,li2022cross}, helping the model to disregard variations across the domains that are irrelevant to the task. Self-training, on the other hand, involves training a student model in a supervised fashion on the target dataset using pseudo-labels~\cite{lee2013pseudo}. To improve the quality of the pseudo-labels, many approaches \cite{cai2019exploring,deng2021unbiased,cao2023contrastive,li2022cross,hoyer2022daformer} use a mean teacher \cite{tarvainen2017mean}, which is an exponential moving average of the student's parameters, to generate these labels during training. Nevertheless, incorrect pseudo-labels remain a significant challenge \cite{zheng2021rectifying,cao2023contrastive,li2022cross}. Furthermore, while UDA has shown substantial progress in monocular tasks, adapting it to multi-view perception remains largely unexplored. 

    In one of the few efforts to apply UDA methods to multi-view pedestrian detection, Lima \etal \cite{lima2023toward} proposed adapting the detector from \cite{gmvd} to unlabeled target data using self-training. However, the method suffered from low-quality pseudo-labels, resulting in modest improvements on a single benchmark. 
    Lima \etal later improved their approached by incorporating a mean teacher for pseudo-labeling \cite{lima2024mean}. However, the success of the method is conditioned on pre-training with pseudo-labels generated by an external detector \cite{lima20223d}, which in turn relies on supervised training on large, labeled datasets for monocular human pose estimation. As a result, the approach still requires substantial amounts of labeled data, which may limit its practical use. 
    In contrast to these methods, our work presents a solution for unsupervised domain adaptation in multi-view pedestrian detection that does not depend on any auxiliary labeled datasets or pre-trained models derived from them.

\section{Methods}
    \begin{figure*}
        \centering
        \includegraphics[width=0.7\linewidth]{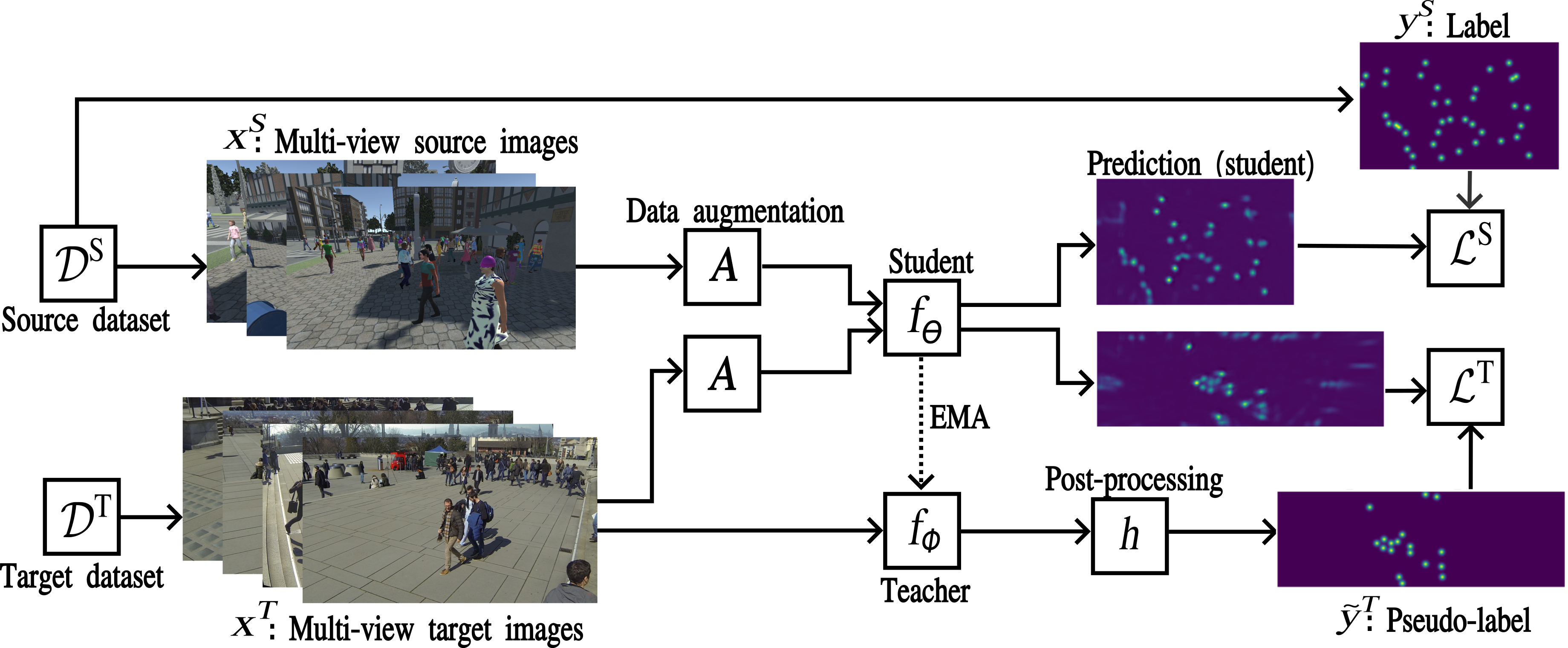}
        \caption{An overview of our proposed self-training method for UDA multi-view pedestrian detection. A student is trained with labels on the source domain and pseudo-labels on the target domain, which are created by a mean teacher. While the teacher creates pseudo-labels on unaugmented data, the student receives strongly augmented images. Note that the label and pseudo-label have been \textit{softened} with a Gaussian kernel in this figure to ease visualization.}
        \label{fig:self-training}
    \end{figure*}
In this section, we introduce our UDA method for multi-view pedestrian detection, designed to leverage labeled source data alongside unlabeled target data to train a multi-view detector for deployment on the target domain. We begin by detailing the detector architecture. Thereafter, we outline our overall UDA strategy and, finally, introduce our approach for generating high-quality pseudo-labels.

\label{sec:method}
\subsection{Multi-view detector}
 
Due to its simplicity and good generalization capability, we use the multi-view detector of \cite{gmvd}, a variant of \cite{mvdet}, which consists of three components: 2D image feature extraction, perspective transformation, and spatial aggregation.

\textbf{Feature extractor:} Given $N$ RGB-images from different views, 
a ResNet-18 \cite{he2016deep} extracts features with $C$ channels and spatial dimension $H_f \times W_f$ for each view.

\textbf{Perspective transformation:} Assuming a known camera calibration matrix for each camera,
the output of the feature extractor are transformed to BEV using a perspective transformation. 
The result of this operation is $N$ BEV feature maps of shape $C \times H_g \times W_g$, where $H_g$ and $W_g$ defines the spatial dimension of the BEV. The purpose is to put all features in the common BEV, which prepares them for spatial aggregation. For a detailed explanation, we refer the reader to the original paper \cite{mvdet}.

\textbf{Spatial aggregation:} The BEV features from different cameras are concatenated to produce a BEV feature map of shape $N \times C \times H_g \times W_g$. Average pooling is then applied along the first dimension to reduce its shape to $C \times H_g \times W_g$. Since average pooling makes the shape of the BEV feature map independent of the number of views $N$, it allows for naturally handling a varying number of cameras. Finally, three dilated convolutional layers process the BEV feature map to regress the probabilistic occupancy map of dimension $H_g \times W_g$. During inference, the probabilistic occupancy map is thresholded to produce detection candidates, which are then subject to non-maximum suppression (NMS) to remove duplicate detections.


\subsection{Mean teacher self-training}
\label{preliminary}
    
     In multi-view detection, a labeled source dataset with $N_s$ samples can be described as $\mathcal{D}^S = \{(x^{S,k}, y^{S,k})\}_{k=1}^{N_S}$, where $x^{S,k}$ denotes a batch of multi-view images from the source domain and $y^{S,k}$ denotes the associated occupancy map label. Similarly, an unlabeled target dataset with $N_T$ samples is described by $\mathcal{D}^T = \{x^{T,k}\}_{k=1}^{N_T}$, where $x^{T,k}$ is a batch for multi-view images from the target domain. 
     In established self-training methods for monocular perception, a model $f_{\theta}$ (the student) is trained on labeled samples from the source dataset and pseudo-labeled samples from the target dataset.
     Note that $f_{\theta}$ in our case is the multi-view detector described in the previous section. 
     Moreover, the pseudo-labels are typically created during training by a mean teacher $f_{\phi}$. The architecture of $f_{\phi}$ is the same as $f_{\theta}$, but its weights $\phi$ are updated as an exponential moving average of the student's weights $\theta$ according to  

     \begin{equation}
     \label{eq:ema}
         \phi_{t+1} \leftarrow \alpha \phi_{t} + (1 - \alpha) \theta_{t},
     \end{equation}
     where $\alpha$ is a hyperparameter.
     Formally, the pseudo-label $\tilde{y}^T$ for a batch of multi-view images $x^T$ on the target domain (dropping the index $k$ for ease of notation) is defined by
         
    \begin{equation}
    \label{eq:pseudo_label}
        \tilde{y}^T = h(f_{\phi}(x^T)) ,
    \end{equation}
    where $h$ denotes the post-processing function that maps the predictions to pseudo-labels. In multi-view pedestrian detection, $h$ typically consist of applying a threshold to the predicted occupancy map and then applying non-maximum suppression. In this work, we consider both conventional post-processing and our own proposal, which will be described in the next section.  
    Furthermore, while $f_{\phi}$ is fed target images $x^T$ for pseudo-labeling, the student is fed augmented images $A(x^T)$. In our work, we also augment the source images $x^S$ to improve the student's generalization capability. Thus, the weights $\theta$ of the student network $f_{\theta}$ are trained to minimize the loss 

    \begin{equation}
     L(\theta) = \mathbb{E}[\mathcal{L}^S(y^{S}, f_{\theta}(A(x^S))) + \lambda \mathcal{L}^T(\tilde{y}^T, f_{\theta}(A(x^T)))],
    \label{uda_loss}
    \end{equation}
    where the expectation is taken over data from the source and target datasets and $\lambda$ is a hyperparameter that adjusts the influence of the target data. Following \cite{mvdet}, we apply a Gaussian kernel $G(\cdot)$ to generate a \textit{soft} target and train the model with the MSE loss. We adopt this loss for both the source and target domain according to        
    \begin{equation}
     \mathcal{L}^S(y, \hat{y}) = \mathcal{L}^T(y, \hat{y}) = 
     \sum_{i=1}^{H_g} \sum_{j=1}^{W_g} (G(y_{ij}) - \hat{y}_{ij})^2,
    \label{mse_loss}
    \end{equation}
    where $y$ and $\hat{y}$ denotes a label (or pseudo-label) and prediction respectively. The proposed mean teacher self-training framework is schematically illustrated in \cref{fig:self-training}. Before adapting the model to the target domain, however, we pre-train it using only source data.

\subsection{Local-max pseudo-labeling}

An essential step in the self-training framework detailed in the previous section is the creation of pseudo-labels. In multi-view pedestrian detection, post-processing is applied to the predicted probabilistic occupancy map to derive a set of detections. In this section, we first review the conventional post-processing method and then introduce our alternative, which is tailored for the UDA problem.

\textbf{Vanilla pseudo-labeling:} The conventional method, adopted by e.g. \cite{mvdet, gmvd, 3drom, hou2021multiview}, comprises the following steps: First, all candidate locations with confidence scores exceeding a threshold $\tau$ are added to a list, sorted in descending order by score. Second, the algorithm selects the first candidate in the list as a detection and removes all candidates within a Euclidean distance $d$ of this detection. Third, the second step is repeated until the list is empty.

\begin{figure}[]
    \centering
    \includegraphics[width=1.0\linewidth]{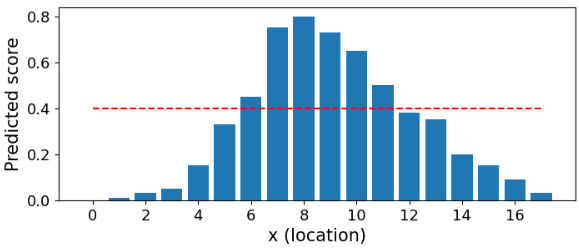}
    \caption{Illustrative example of predicted occupancy scores in one dimension.}
    \label{fig:vanilla-pseudo}
\end{figure}

To illustrate, consider a one-dimensional example with $\tau=0.4$ and $d=2$, shown in Figure \ref{fig:vanilla-pseudo}. Here, six candidates on positions $x\in \{6,7,8,9,10, 11\}$ exceed the threshold and are added to the list. Since position $x=8$ has the highest confidence, it is selected as the first detection. Subsequently, candidates at positions 6,7,9, and 10 are removed from the list because they fall within distance $d$ of the first detection. The candidate at position 11 is then selected as a second detection. The algorithm terminates at this point since no candidates remain in the list. Note, however, that if the threshold $\tau$ had been lower, a third detection at, e.g., $x=5$ could have been attained.  

Since the predicted confidence level on the target domain is difficult to foresee, we question whether this post-processing method is overly dependent on the threshold $\tau$. Ideally, a well trained network is expected to exhibit predictions with a single local maxima at each location of a pedestrian following training with the MSE loss on the Gaussian targets described in \cref{mse_loss}. However, this post-processing method may also produce detections that are not local maxima. We hypothesize that such detections are less reliable, especially in UDA when the threshold $\tau$ is ambiguous.   

\textbf{Local-max pseudo-labeling:} Motivated by the above analysis, we propose an alternative post-processing method that only considers points that are \textit{local maxima} as candidate detections. To allow for an efficient implementation, in for example PyTorch, we define a local maxima as a position $ij$ in the occupancy map for which the predicted confidence $\hat{y}_{ij}$ satisfies 
\begin{equation}
    \hat{y}_{ij} \geq \hat{y}_{kl} \text{ } \forall k \in [i-k_d, i+k_d] \text{ and } \forall l \in [j-k_d, j+k_d],
\label{eq:local-max}
\end{equation}
where $k$ and $l$ are integers and $k_d$ is a parameter that defines the size of the considered neighborhood. Since the predictions are expected to exhibit some degree of noise, we also require the predicted confidence of any detections to exceed the threshold $\tau$. Note, however, that a location that is not a local maxima is never considered a candidate detection in our method, regardless of the value of $\tau$, which distinguishes it from the conventional method.

\section{Experiments}

    \subsection{Experimental setup}
        \textbf{Datasets:} we use the popular Wildtrack \cite{chavdarova2018wildtrack} and MultiviewX \cite{mvdet} datasets as well as a subset of the newly introduced GMVD \cite{gmvd} dataset. Wildtrack is a real-world dataset comprising 400 multi-view images collected from a single camera rig of seven cameras with overlapping fields of view, covering an area of 12x36 meters. For annotation, the ground plane is discretized into a 480x1440 grid, where each cell corresponds to a $2.5$x$2.5$ cm region. Meanwhile, MultiviewX is a synthetic dataset of 400 images from six cameras covering an area of 16x25 meters, with a grid shape of 640x1000 of the same spatial resolution. GMVD is another synthetic dataset, distinct for its multiple scenes and camera configurations. Here, the covered area depends on the scene and the grid is chosen to attain the same spatial resolution of $2.5$x$2.5$ cm.

        We consider the benchmark MultiviewX$\rightarrow$Wildtrack to evaluate adaptation from labeled simulated data to unlabeled real-world data, and the converse, which we denote as Wildtrack$\rightarrow$MultiviewX. Following \cite{gmvd}, we also consider the intra-dataset benchmarks Wildtrack 1,3,5,7$\rightarrow$2,4,5,6, Wildtrack 2,4,5,6$\rightarrow$1,3,5,7, and MultiviewX 1,2,6$\rightarrow$3,4,5, where different subset of cameras from a single dataset constitute the source and target domain. The purpose is to evaluate adaptation across camera rigs without the presence of a sim-to-real domain gap. Additionally, to address this problem in the more challenging setting where the source and target datasets are collected from different scenes, we introduce two new benchmarks wherein GMVD and MultiviewX constitute the source and target domain respectively. Like the intra-dataset benchmarks, we consider labels on a single camera rig and therefore use only a subset of GMVD as the labeled source dataset. Specifically, we consider two different camera configurations on the first scene of GMVD as the source domain and introduce the benchmarks GMVD1$\rightarrow$MultiviewX and GMVD2$\rightarrow$MultiviewX. For all benchmarks, we use the first 90\% of samples in MultiviewX and Wildtrack for training and the last 10\% for testing. GMVD1 and GMVD2 both consists of five cameras and comprise 517 training frames. 

        \textbf{Evalation metrics:} like most previous works, we evaluate the models in terms of the MODA, MODP, precision and recall metrics. MODA serves as the primary performance indicator, since it accounts for both missed detections and false positives, while MODP evaluates the localization precision \cite{kasturi2008framework}. For all metrics, we report the performance in percentage.
        
    \subsection{Implementation details}
        Following \cite{gmvd}, input images are resized to shape 720x1280 before being processed by ResNet-18 \cite{he2016deep}, extracting 512-channel feature maps. These features are resized to shape 270x480 through bilinear interpolation before being projected to the ground plane, whose shape depends on the dataset. 
        For spatial aggregation, we employ three convolutional layers with kernel size 3 and dilation factors of 1, 2 and 4. For training, we use the one-cycle learning rate scheduler \cite{smith2019super} with a max learning rate of 0.1 and the SGD optimizer with momentum 0.5 and L2 regularization $5\cdot 10^{-4}$. We use a batch size of 1 and employ early stopping to avoid overfitting. For evaluation, we use (conventional) NMS with a spatial threshold of 0.5 meters like previous works \cite{mvdet, gmvd}. 
        However, while these works use the threshold $\tau=0.4$, we evaluate the model on the range $\tau \in \{0.05, 0.10, ..., 0.95\}$ and select the result with highest MODA. The purpose is to ensure that the experimental results are not affected by the specific choice of $\tau$, whose optimal value is ambiguous in the UDA setting. 
        
        Prior to self-training, we initialize ResNet-18 with ImageNet \cite{deng2009imagenet} weights and pre-train the model on only source data for 20 epochs, which constitutes our \textit{Baseline}. Unless stated otherwise, the UDA results are obtained by adapting the baseline to the target domain by 5 epochs of self-training, using $\lambda=1.0$, $\alpha=0.99$, and the proposed local-max pseudo-labeling with $k_d=3$. The threshold $\tau$ is experimentally set to $0.4$ for MultiviewX$\rightarrow$Wildtrack, $0.2$ for Wildtrack$\rightarrow$MultiviewX, and $0.3$ for all other benchmarks, which is motivated in \cref{sec:results-in-depth}. Moreover, Dropview \cite{gmvd} and 3DROM \cite{3drom} augmentation is used both to train the baseline and in self-training.        
    
    \subsection{MVUDA compared with previous methods}
        In this section, we compare our UDA method with previous SOTA methods, as well as our \textit{Baseline} (trained only on source), and the \textit{Oracle}, which is trained with labels on the target domain similarly as the baseline was trained on the source domain. For qualitative results, please refer to the supplementary material. In \cref{tab:main-res}, the results on MultiviewX$\rightarrow$Wildtrack and Wildtrack$\rightarrow$MultiviewX are presented. The dashed line separates the methods that use auxiliary labeled datasets from those that use labels only on the source domain.  
        It can be seen that our UDA method boosts the baseline performance significantly with respect to all studied metrics on both benchmarks. Our UDA method also achieves the highest MODA among the methods that don't rely on auxiliary labeled data. Impressively, our UDA method boosts the baseline from 35.9 to 82.4 MODA on Wildtrack$\rightarrow$MultiviewX, outperforming \cite{lima2024mean} by a large margin although they rely on a monocular detector derived from large, labeled monocular datasets.

        In \cref{tab:main-res-uncommon}, we further evaluate our method on five camera rig adaptation benchmarks. In all cases, our UDA method significantly boosts the baseline in terms of MODA. We also outperform \cite{gmvd} on the two Wildtrack benchmarks proposed by them. Furthermore, our UDA method reaches close to Oracle performance on the two GMVD$\rightarrow$MultiviewX benchmarks. 
        Interestingly, the gap between our UDA method and the Oracle is slightly higher for the three intra-dataset benchmarks, suggesting that our method is less effective when the number of cameras is smaller.
        It is worth mentioning that we don't compare our results to \cite{gmvd} on MultiviewX 1,2,6$\rightarrow$3,4,5 because they use a different evaluation protocol, evaluating only on a subset of the labels while we use all labels. 
         
\begin{table}[]
\centering
\tabcolsep=4.5pt
\begin{tabular}{lllll}
\hline
\multicolumn{1}{l|}{Method}       & MODA & MODP & Precision & Recall \\ \hline
\multicolumn{5}{c}{MultiviewX $\rightarrow$ Wildtrack}               \\ \hline
\multicolumn{1}{l|}{$^{\dagger}$Lima et al. \cite{lima2024mean}}  & 85.1 & 74.8 & 93.9      & 91.0   \\
\multicolumn{1}{l|}{$^{\dagger}$PPM \cite{qiu2024ppm}}          & 90.3 & 72.6 & 94.4      & 96.0   \\ 
\multicolumn{1}{l|}{Oracle}       & 91.3 & 75.5 & 97.0      & 94.2   \\ \hdashline
\multicolumn{1}{l|}{GMVD \cite{gmvd}}         & 70.7 & 73.8 & 89.1      & 80.6   \\
\multicolumn{1}{l|}{TMVD \cite{deeptopdown}}         & 74.9 & \textbf{76.9} & 90.4      & 83.8   \\
\multicolumn{1}{l|}{MVFP \cite{mvfp}}         & 82.6 & 76.2 & 89.6      & \textbf{93.4}   \\
\multicolumn{1}{l|}{Baseline} & 70.0 & 73.6 & 89.2      & 79.6   \\
\multicolumn{1}{l|}{MVUDA (ours)}      & \textbf{85.4} & 75.3 & \textbf{96.5}      & 88.7   \\ \hline
\multicolumn{5}{c}{Wildtrack $\rightarrow${}MultiviewX}              \\ \hline
\multicolumn{1}{l|}{$^{\dagger}$Lima et al. \cite{lima2024mean}}  & 75.9 & 78.6 & 96.2      & 79.0   \\
\multicolumn{1}{l|}{Oracle}       & 91.2 & 82.1 & 97.5      & 93.6   \\ \hdashline
\multicolumn{1}{l|}{Baseline} & 35.9 & 66.4 & 82.8      & 45.2   \\
\multicolumn{1}{l|}{MVUDA (ours)}      & \textbf{82.4} & \textbf{75.4} & \textbf{93.3}      & \textbf{88.8}  
\end{tabular}
\caption{Performance comparison with state-of-the-art methods on two cross-domain benchmarks. The methods marked with $^{\dagger}$ rely on models trained on large, labeled datasets for monocular vision.}
\label{tab:main-res}
\end{table}

\begin{table}[]
\centering
\tabcolsep=4.5pt
\begin{tabular}{lllll}
\hline
\multicolumn{1}{l|}{Method}        & MODA & MODP & Precision & Recall \\ \hline

\multicolumn{5}{c}{Wildtrack 2,4,5,6 $\rightarrow$ 1,3,5,7}           \\ \hline
\multicolumn{1}{l|}{Oracle} & 83.7 & 75.8 & 94.6      & 88.8   \\ \hdashline
\multicolumn{1}{l|}{GMVD \cite{gmvd}}          & 75.1 & 71.1 & 94.3      & 79.9   \\
\multicolumn{1}{l|}{Baseline}  & 75.2 & 71.1 & 91.5      & \textbf{82.9}   \\
\multicolumn{1}{l|}{MVUDA (ours)}       & \textbf{79.4} & \textbf{77.8} & \textbf{96.3}      & 82.6   \\ \hline

\multicolumn{5}{c}{Wildtrack 1,3,5,7 $\rightarrow$ 2,4,5,6}           \\ \hline
\multicolumn{1}{l|}{Oracle}        & 87.3 & 71.4 & 94.5      & 92.6   \\ \hdashline
\multicolumn{1}{l|}{GMVD \cite{gmvd}}          & 62.6 & 67.4 & 86.7      & 73.9   \\
\multicolumn{1}{l|}{Baseline}  & 72.3 & 68.1 & 88.1      & 83.5   \\
\multicolumn{1}{l|}{MVUDA (ours)}       & \textbf{81.4} & \textbf{68.8} & \textbf{95.9}      & \textbf{85.1}   \\ \hline

\multicolumn{5}{c}{MultiviewX 1,2,6$\rightarrow${}3,4,5}              \\ \hline
\multicolumn{1}{l|}{Oracle}        & 75.6 & 74.1 & 95.3      & 79.5 \\ \hdashline
\multicolumn{1}{l|}{Baseline}  & 54.7 & 69.0 & 89.8      & 61.7   \\
\multicolumn{1}{l|}{MVUDA (ours)}       & \textbf{64.2} & \textbf{73.0} & \textbf{91.3}      & \textbf{71.0} \\ \hline 

\multicolumn{5}{c}{GMVD1 $\rightarrow$ MultiviewX}             \\ \hline
\multicolumn{1}{l|}{Oracle}        & 91.2 & 82.1 & 97.5      & 93.6   \\ \hdashline
\multicolumn{1}{l|}{Baseline}  & 70.3 & 74.5 & 89.7      & 79.5   \\
\multicolumn{1}{l|}{MVUDA (ours)}       & \textbf{89.0} & \textbf{78.4} & \textbf{97.0}      & \textbf{91.8}   \\ \hline

\multicolumn{5}{c}{GMVD2 $\rightarrow$ MultiviewX}             \\ \hline
\multicolumn{1}{l|}{Oracle}        & 91.2 & 82.1 & 97.5      & 93.6   \\ \hdashline
\multicolumn{1}{l|}{Baseline}  & 66.9 & 74.0 & 85.8      & 80.1   \\
\multicolumn{1}{l|}{MVUDA (ours)}       & \textbf{88.8} & \textbf{76.9} & \textbf{97.2}      & \textbf{91.5}   

\end{tabular}
\caption{Performance comparison with state-of-the-art methods on five different camera rig adaptation benchmarks.}
\label{tab:main-res-uncommon}
\end{table}

\subsection{Ablation study}
    To study the importance of Mean Teacher (MT) and data augmentation (Aug) in the self-training (ST) framework, we ablate these components on two benchmarks in \cref{tab:ablation}. 
    Here, the first row shows the performance without any adaptation (baseline). Furthermore, self-training without mean teacher implies that the (frozen) baseline model creates pseudo-labels throughout training. It can be seen that self-training alone yields substantial improvements over the baseline.  Moreover, the results improves significantly when adding the mean teacher and the data augmentation. It is noteworthy that the impact of data augmentation is greater on the sim-to-real benchmark, where it may serve as key component in overcoming the larger domain gap. 
    
\begin{table}[h!]
\centering
\tabcolsep=4.5pt
\begin{tabular}{lllllll}
\hline
ST & MT & \multicolumn{1}{l|}{Aug} & MODA & MODP & Precision & Recall \\ \hline
\multicolumn{7}{c}{MultiviewX $\rightarrow$ Wildtrack}                \\ \hline
   &    & \multicolumn{1}{l|}{}    & 70.0 & 73.6 & 89.2      & 79.6   \\
$\checkmark$  &    & \multicolumn{1}{l|}{}    & 75.0 & 73.3 & 92.0      & 82.1   \\
$\checkmark$  & $\checkmark$  & \multicolumn{1}{l|}{}    & 78.7 & 74.2 & 92.1      & 86.0   \\
$\checkmark$  & $\checkmark$  & \multicolumn{1}{l|}{$\checkmark$}   & \textbf{85.4} & \textbf{75.3} & \textbf{96.5}      & \textbf{88.7}   \\ \hline

\multicolumn{7}{c}{GMVD1 $\rightarrow${}MultiviewX}                   \\ \hline
   &    & \multicolumn{1}{l|}{}    & 70.3 & 74.5 & 89.7      & 79.5   \\
$\checkmark$  &    & \multicolumn{1}{l|}{}    & 76.6 & 76.0 & 91.5      & 84.5   \\
$\checkmark$  & $\checkmark$  & \multicolumn{1}{l|}{}    & 87.2 & 76.6 & \textbf{97.6}      & 89.4   \\
$\checkmark$  & $\checkmark$  & \multicolumn{1}{l|}{$\checkmark$}   & \textbf{89.0} & \textbf{78.4} & 97.0      & \textbf{91.8}  
\end{tabular}
    \caption{Ablation of the Mean Teacher (MT) and data augmentation (Aug), which are two pivotal components in the self-training (ST) framework.}
        \label{tab:ablation}
    \end{table}

    \subsection{In-depth analysis of MVUDA}
    \label{sec:results-in-depth}

    In this section, we analyze key components of our proposed method in detail, including the introduced pseudo-labeling technique, the parameter $\alpha$, and the data augmentation. Unless stated otherwise, herein self-training comprises local-max pseudo-labeling with $k_d=3$, $\alpha=0.99$, $\lambda=1$, and no data augmentation. Again, the threshold $\tau$ is set to $0.4$ for MultiviewX$\rightarrow$Wildtrack, $0.2$ for Wildtrack$\rightarrow$MultiviewX, and $0.3$ for all other benchmarks, following the experiments presented in \cref{tab:vanilla_vs_max}.  
    
        \textbf{Pseudo-labeling:} 
        Table \ref{tab:vanilla_vs_max} shows MODA of our UDA method using either vanilla pseudo-labeling or local-max pseudo-labeling. For convenience, we show the MODA of the baseline (from \cref{tab:main-res,tab:main-res-uncommon}) in parenthesis in the benchmark headings. Missing values mean that no improvement over the baseline was obtained. It can be seen that the best performance is achieved using our pseudo-labeling method on all benchmarks except the first one, where the vanilla method performs slightly better. Notably, our method outperforms the vanilla method by more than 25 MODA on Wildtrack$\rightarrow$MultiviewX. Furthermore, the proposed method yields improvements over the baseline for a wider range of $\tau$, demonstrating improved robustness to the choice of this hyperparameter.

\begin{table}[]
\centering
\tabcolsep=4.5pt
\begin{tabular}{lrlll}
\multicolumn{1}{l|}{Method}        & \multicolumn{1}{l}{$\tau=$0.2} & 0.3  & 0.4  & 0.5  \\ \hline
\multicolumn{5}{c}{MultiviewX $\rightarrow$ Wildtrack (70.0)}                            \\ \hline
\multicolumn{1}{l|}{UDA vanilla}   & -                              & -    & \textbf{78.6} & 72.2 \\
\multicolumn{1}{l|}{UDA local-max} & -                              & 70.8 & 75.8 & -    \\ \hline
\multicolumn{5}{c}{Wildtrack $\rightarrow$ MultiviewX (35.9)}                            \\ \hline
\multicolumn{1}{l|}{UDA vanilla}   & -                              & 48.1 & 47.9 & -    \\
\multicolumn{1}{l|}{UDA local-max} & \textbf{73.2}                           & 68.7 & 43.5 & -    \\ \hline
\multicolumn{5}{c}{Wildtrack 2,4,5,6 $\rightarrow$ 1,3,5,7 (75.2)}                       \\ \hline
\multicolumn{1}{l|}{UDA vanilla}   & -                              & -    & 78.5 & -    \\
\multicolumn{1}{l|}{UDA local-max} & -                              & \textbf{78.6} & 77.7 & -    \\ \hline
\multicolumn{5}{c}{Wildtrack 1,3,5,7 $\rightarrow$ 2,4,5,6 (72.3)}                       \\ \hline
\multicolumn{1}{l|}{UDA vanilla}   & -                              & -    & 73.4 & -    \\
\multicolumn{1}{l|}{UDA local-max} & -                              & \textbf{79.8} & -    & -    \\ \hline
\multicolumn{5}{c}{MultiviewX 1,2,3$\rightarrow${}4,5,6 (54.7)}                          \\ \hline
\multicolumn{1}{l|}{UDA vanilla}   & -                              & -    & 55.2 & -    \\
\multicolumn{1}{l|}{UDA local-max} & 58.1                           & \textbf{63.1} & 56.3 & -   \\ \hline

\multicolumn{5}{c}{GMVD1 $\rightarrow$ MultiviewX (70.3)}                                \\ \hline
\multicolumn{1}{l|}{UDA vanilla}   & -                              & \textbf{87.8} & 81.5 & -    \\
\multicolumn{1}{l|}{UDA local-max} & 73.4                           & \textbf{87.8} & 81.3 & -    \\ \hline
\multicolumn{5}{c}{GMVD2 $\rightarrow$ MultiviewX (66.9)}                                \\ \hline
\multicolumn{1}{l|}{UDA vanilla}   & -                              & 74.9 & 82.8 & -    \\
\multicolumn{1}{l|}{UDA local-max} & 79.9                           & \textbf{88.1} & 80.1 & -    

\end{tabular}
\caption{Performance comparison (MODA) of self-training with vanilla or local-max pseudo-labeling at different thresholds $\tau$.}
\label{tab:vanilla_vs_max}
\end{table}

To understand these results, we analyze the performance of the baseline model when evaluated using either of the two post-processing methods. \Cref{tab:post-processing} shows the results on MultiviewX$\rightarrow$Wildtrack and Wildtrack$\rightarrow$MultiviewX for different thresholds $\tau$. It can be seen that our post-processing method attains higher precision and MODP in all cases, demonstrating that detections that are local maxima are typically more reliable. However, recall is higher for the vanilla method, owing to the fact that it usually produces a larger number of detections. It is noteworthy that the difference between the two methods is more pronounced for small values of $\tau$. This is because vanilla post-processing produces many detections that are not local maxima in this case. Since these detections are less reliable, our method attains much higher MODA in this regime. Consequently, our method is able to harness reliable pseudo-labels at lower confidence levels, which evidently is particularly beneficial on the Wildtrack$\rightarrow$MultiviewX benchmark.    

\begin{table*}[]
\centering
\tabcolsep=4.5pt
\begin{tabular}{crlllllllllllllll}
\hline
\multicolumn{1}{l|}{}        & \multicolumn{4}{c|}{MODA}                                                 & \multicolumn{4}{c|}{MODP}                      & \multicolumn{4}{c|}{Precision}                 & \multicolumn{4}{c}{Recall} \\ \hline
\multicolumn{1}{l|}{Method}  & \multicolumn{1}{l}{$\tau=$ 0.2} & 0.3  & 0.4  & \multicolumn{1}{l|}{0.5}  & 0.2  & 0.3  & 0.4  & \multicolumn{1}{l|}{0.5}  & 0.2  & 0.3  & 0.4  & \multicolumn{1}{l|}{0.5}  & 0.2   & 0.3  & 0.4  & 0.5  \\ \hline
\multicolumn{17}{c}{MultiviewX $\rightarrow$ Wildtrack}                                                                                                                                                                                 \\ \hline
\multicolumn{1}{l|}{Vanilla} & 0.0                             & 42.9 & 70.0 & \multicolumn{1}{l|}{\textbf{63.1}} & 71.9 & 72.7 & 73.6 & \multicolumn{1}{l|}{75.0} & 40.0 & 66.0 & 89.2 & \multicolumn{1}{l|}{97.3} & \textbf{95.5}  & \textbf{88.2} & \textbf{79.6} & \textbf{64.9} \\
\multicolumn{1}{l|}{Local-max}    & \textbf{42.0}                            & \textbf{59.2} & \textbf{70.1} & \multicolumn{1}{l|}{62.6} & \textbf{72.4} & \textbf{73.2} & \textbf{73.9} & \multicolumn{1}{l|}{\textbf{75.1}} & \textbf{65.9} & \textbf{78.1} & \textbf{92.4} & \multicolumn{1}{l|}{\textbf{97.6}} & 87.0  & 82.4 & 76.4 & 64.2 \\ \hline
\multicolumn{17}{c}{Wildtrack $\rightarrow${}MultiviewX}                                                                                                                                                                                \\ \hline
\multicolumn{1}{l|}{Vanilla} & 25.0                            & 35.9 & 32.5 & \multicolumn{1}{l|}{24.7} & 64.2 & 66.4 & 67.1 & \multicolumn{1}{l|}{68.6} & 63.9 & 82.8 & 92.6 & \multicolumn{1}{l|}{95.6} & \textbf{57.2}  & \textbf{45.2} & \textbf{35.3} & \textbf{25.9} \\
\multicolumn{1}{l|}{Local-max}    & \textbf{48.5}                            & \textbf{41.5} & \textbf{33.2} & \multicolumn{1}{l|}{\textbf{24.8}} & \textbf{66.0} & \textbf{67.5} & \textbf{68.2} & \multicolumn{1}{l|}{\textbf{69.3}} & \textbf{95.1} & \textbf{98.6} & \textbf{99.0} & \multicolumn{1}{l|}{\textbf{98.9}} & 51.1  & 42.1 & 33.5 & 25.1
\end{tabular}
\caption{Performance of the baseline when evaluated using either vanilla or the proposed local-max post-processing.}
\label{tab:post-processing}
\end{table*}
To further validate the robustness of our method, we analyze the performance using different values of the parameter $k_d$ on two benchmarks.
It is noteworthy that the considered neighborhood for local-max pseudo-labeling, defined in \cref{eq:local-max}, is a square of size 70x70 cm when $k_d=3$ since each cell in the predicted occupancy map corresponds to 10x10 cm. Hence, $k_d=3$ is the largest value for which the entire square is within a radius of 0.5 meters, which is the distance used in NMS by conventional methods. In \cref{tab:kernel_size}, it can be seen that our method works well as long as $k_d$ is sufficiently small. Interestingly, the method works well even with the smallest possible value of $k_d=1$. One could perhaps expect that the method would produce many false positives due to noise in the predictions with such a small kernel. Conversely, the predictions exhibit a reasonable smoothness that mitigates this problem, adding to the robustness of our method. Since $k_d$ acts as a lower bound on the distance between any two pseudo-labels, a too large $k_d$ risks degrading performance since it may introduce false negatives in crowded scenes, which happens around $k_d = 7$ on MultiviewX$\rightarrow$Wildtrack. 

\begin{table}[]
\begin{tabular}{llllll}
\hline
$k_d=$   & 1      & 2      & 3      & 5     & 7    \\ \hline
\multicolumn{6}{c}{MultiviewX $\rightarrow$ Wildtrack (70.0)} \\ \hline
                  & \textbf{81.2}   & 80.9   & 79.9   & 78.3   & 63.4  \\ \hline
\multicolumn{6}{c}{GMVD1 $\rightarrow${}MultiviewX (70.3)}    \\ \hline
                  & 86.9   & \textbf{88.1}   & 88.0   & 87.8   & 86.4 
\end{tabular}
\caption{Performance comparison (MODA) of self-training using local-max pseudo-labeling with different values of $k_d$.}
\label{tab:kernel_size}
\end{table}

\textbf{Mean teacher $\alpha$:} 
Table \ref{tab:alpha_teacher} show the performance in MODA of our UDA method when trained for either 5 or 20 epochs with different values of the parameter $\alpha$. Note that $\alpha=0$ implies that the teacher model equals the student (i.e., the student model is creating pseudo-labels), while $\alpha=1$ implies that the frozen baseline model creates the pseudo-labels throughout training. It can be seen that both $\alpha=0.99$ and $\alpha=0.999$ yields decent performance on both benchmarks when training for 5 and 20 epochs, although a slowly evolving teacher ($\alpha=0.999$) seems to benefit from longer trainings. We also note that a too low value of $\alpha$ leads to stability issues on one benchmark, owing to the rapid updates of the teacher model. Moreover, while freezing the teacher with $\alpha=1$ works reasonable well for both benchmarks, it doesn't yield the best performance since it misses the opportunity to improve the quality of the pseudo-labels as training progresses. For additional experiments, please refer to the supplementary material. 

\begin{table}[]
\tabcolsep=4.5pt
\begin{tabular}{cllllll}
\hline
\multicolumn{1}{l|}{Epochs} & $\alpha=$ & 0    & 0.9  & 0.99 & 0.999 & 1    \\ \hline
\multicolumn{7}{c}{MultiviewX $\rightarrow$ Wildtrack (70.0)}               \\ \hline
\multicolumn{1}{c|}{5}      &           & -    & -    & 79.7 & 76.3  & 77.2 \\
\multicolumn{1}{c|}{20}     &           & -    & -    & 79.1 & \textbf{81.2}  & 77.3 \\ \hline
\multicolumn{7}{c}{GMVD1 $\rightarrow${}MultiviewX (70.3)}                  \\ \hline
\multicolumn{1}{c|}{5}      &           & 85.3 & 88.0 & \textbf{88.2} & 83.5  & 79.0 \\
\multicolumn{1}{c|}{20}     &           & 86.8 & 87.9 & 87.8 & 85.3  & 79.2
\end{tabular}
\caption{Performance comparison (MODA) of self-training for 5 or 20 epochs using different values for $\alpha$.}
\label{tab:alpha_teacher}
\end{table}

\textbf{Data augmentation} Since data augmentation is an essential ingredient in self-training, we investigate three different methods that recently have been proposed for multi-view pedestrian detection. In \cref{tab:data-aug}, we present experiments with Dropview (DV) \cite{gmvd}, 3D random occlusion (3DR) \cite{3drom}, and the two-level data augmentation developed in MVAug (MVA) \cite{mvaug}. 
It can be seen that each of these augmentation methods increases performance on most benchmarks. However, when combining the different methods, the best performance is achieved by DV and 3DR (excluding MVA). Similar results were obtained when we studied the generalization capability of the baseline, for which experiments are provided in the supplementary material. Given the good performance of MVAug presented by \cite{mvaug}, these results are a bit surprising. However, it is also convenient since MVAug is substantially more complex than the other two methods. This is because MVAug, unlike Dropview and 3DR, not only augments the input image, but also augments the perspective transformation applied to the features.  

\begin{table}[]
\begin{tabular}{llllll}
\hline
w/o       & DV        & MVA      & 3DR      & All      & DV+3DR    \\ \hline
\multicolumn{6}{c}{MultiviewX $\rightarrow$ Wildtrack (70.0)}      \\
76.8      & 79.7      & 80.8     & \textbf{85.0}     & 81.8     & 84.7      \\ \hline
\multicolumn{6}{c}{Wildtrack $\rightarrow$ MultiviewX (35.9)}      \\
73.1      & 77.4      & 76.0     & 79.8     & 80.7     & \textbf{82.4}      \\ \hline
\multicolumn{6}{c}{Wildtrack 2,4,5,6 $\rightarrow$ 1,3,5,7 (75.2)} \\
78.0      & 79.3      & \textbf{79.4}     & 79.2     & 79.0     & \textbf{79.4 }     \\ \hline
\multicolumn{6}{c}{Wildtrack 1,3,5,7 $\rightarrow$ 2,4,5,6 (72.3)} \\
79.9      & \textbf{81.9}      & 80.6     & 79.5     & 80.0     & 81.4      \\ \hline
\multicolumn{6}{c}{MultiviewX 1,2,6$\rightarrow${}3,4,5 (54.7)}    \\
62.9      & 63.6      & \textbf{65.1}     & 63.3     & 62.6     & 64.2 \\ \hline 
\multicolumn{6}{c}{GMVD1 $\rightarrow$ MultiviewX (70.3)}          \\
88.0      & 88.3      & 87.1     & 88.8     & 87.0     & \textbf{89.0}      \\ \hline
\multicolumn{6}{c}{GMVD2 $\rightarrow$ MultiviewX (66.9)}   \\
87.9      & 87.8      & 87.7     & \textbf{89.1}     & 87.4     & 88.8         
\end{tabular}
\caption{Performance comparison (MODA) of self-training using different combinations of data augmentation.}
\label{tab:data-aug}
\end{table}

\section{Conclusions}
In this paper, we presented MVUDA, the first unsupervised domain adaptive (UDA) method for multi-view pedestrian detection that eliminates the need for auxiliary labeled datasets. Our approach leverages mean teacher self-training with a novel pseudo-labeling method tailored for multi-view detection, significantly increasing pseudo-label reliability and the effectiveness of the overall framework. Extensive experiments demonstrate the efficacy of our method and motivate key design choices. 
By reducing the reliance on labeled data and achieving superior performance, we believe MVUDA sets a strong baseline for future research in unsupervised domain adaptation and holds significant potential for real-world applications.

\small
\textbf{Acknowledgments} This work is supported by AB Volvo, the Wallenberg AI, Autonomous Systems and Software Program (WASP) funded by the Knut and Alice Wallenberg Foundation, and the Vinnova funded project SMILE IV (2023-00789).
The experiments were enabled by resources provided by the National Academic Infrastructure for Supercomputing in Sweden (NAISS), partially funded by the Swedish Research Council through grant agreement no. 2022-06725.

{
    \small
    \bibliographystyle{ieeenat_fullname}
    \bibliography{main}
}



\clearpage
\setcounter{page}{1}
\maketitlesupplementary

The supplementary material provides qualitative results of our method along with additional experiments. Like in the main paper, the pseudo-labeling threshold $\tau$ for self-training is set to $0.4$ for MultiviewX$\rightarrow$Wildtrack, $0.2$ for Wildtrack$\rightarrow$MultiviewX, and $0.3$ for all other benchmarks, unless stated otherwise. 

\subsection{Qualitative examples}
\begin{figure*}[]
        \centering
        \includegraphics[width=1.0\linewidth]{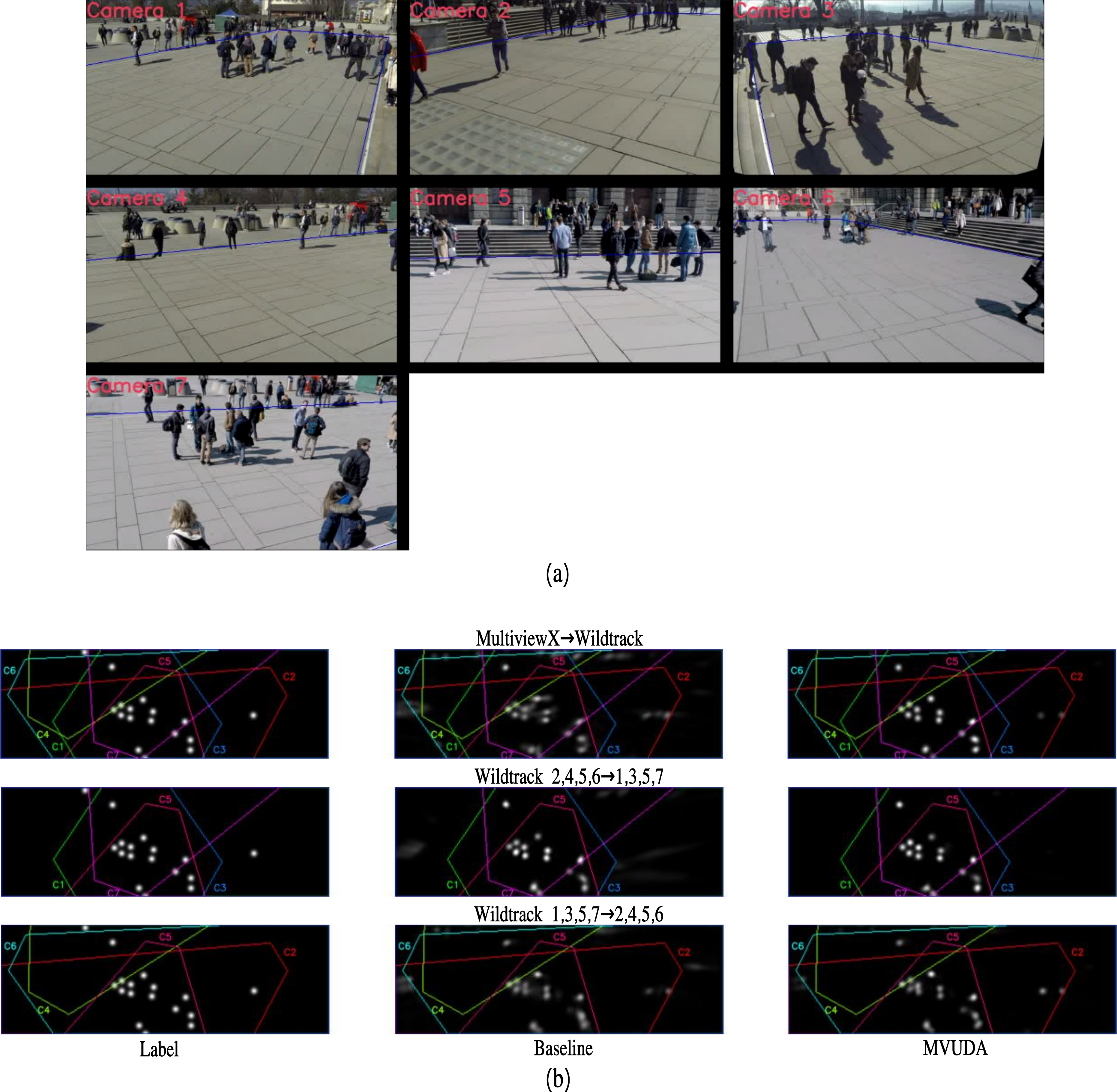}
        \caption{A test sample from Wildtrack (a), as well as the associated label and predictions of the baseline and MVUDA (b). The predictions are produced by the methods trained on the specified benchmark, hence the results differ between the rows. Note that the label is identical across all rows since it is associated with the same test sample (a) in all benchmarks, although only a subset of the available cameras is used in the cases Wildtrack 2,4,5,6$\rightarrow$1,3,5,7 and Wildtrack 1,3,5,7$\rightarrow$2,4,5,6.}
        \label{fig:wildtrack-qualitative}
    \end{figure*}

\begin{figure*}[]
        \centering
        \includegraphics[width=0.8\linewidth]{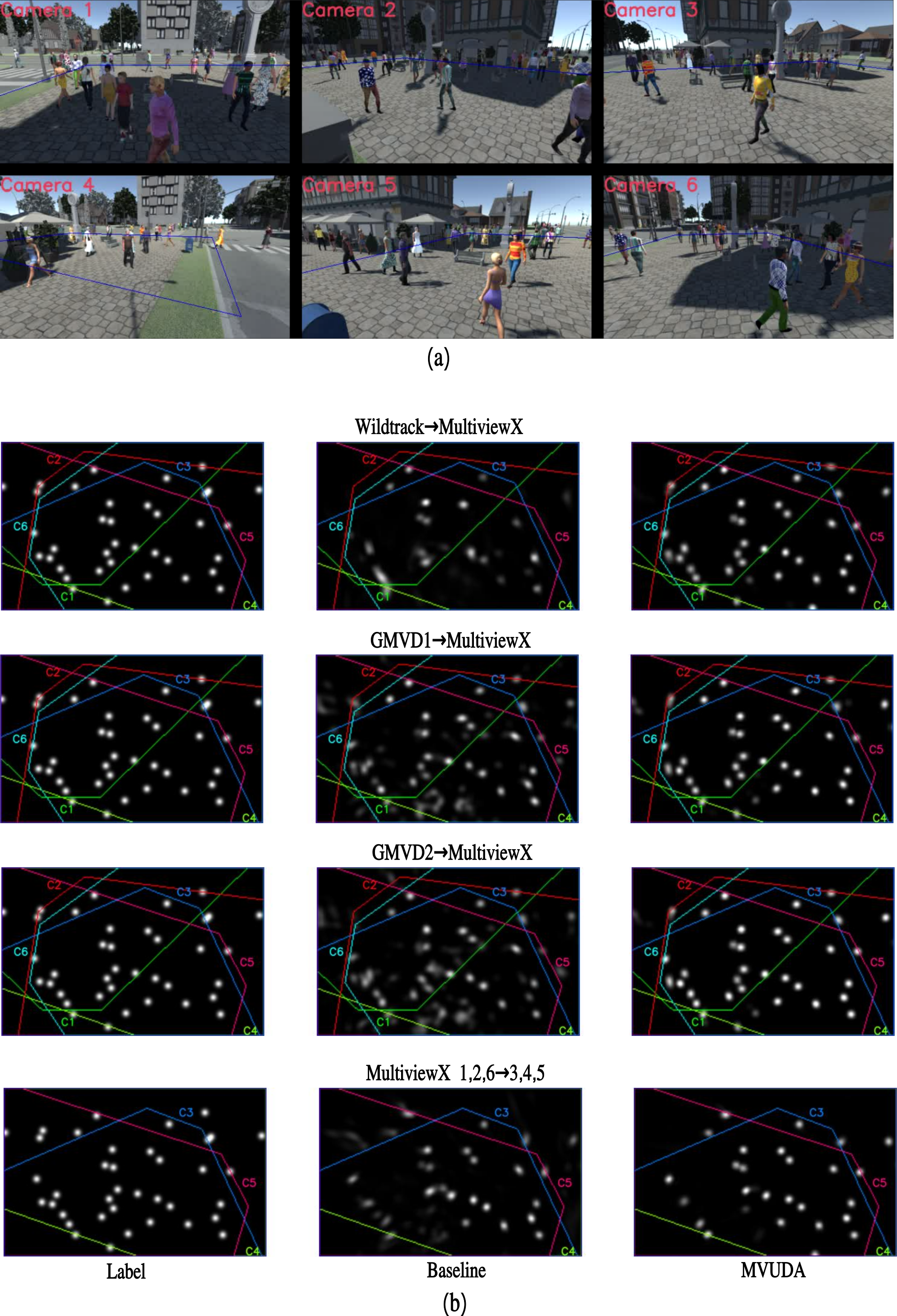}
        \caption{A test sample from MultiViewX (a), as well as the associated label and predictions of the baseline and MVUDA (b). The predictions are produced by the methods trained on the specified benchmark, hence the results differ between the rows. Note that the label is identical across all rows since it is associated with the same test sample (a) in all benchmarks, although only cameras 3,4,5 are used for testing in MultiviewX 1,2,6$\rightarrow$3,4,5.}
        \label{fig:multiviewx-qualitative}
    \end{figure*}

    The predictions of the baseline and MVUDA, quantitatively evaluated in \cref{tab:main-res,tab:main-res-uncommon} of the main paper, are studied qualitatively in this part of the paper. \Cref{fig:wildtrack-qualitative} shows a test sample from the Wildtrack dataset and the associated label and predictions produced by the baseline and MVUDA for different benchmarks. To ease comparison, we visualize the raw predictions (before any post-processing) and the label after \textit{softened} with a Gaussian kernel. It can be seen that MVUDA improves on the baseline mainly in two aspects: first, by reducing the predicted scores in regions where there are no pedestrians, and second, by producing more distinct detections that are not smeared out spatially. Similarly, \cref{fig:multiviewx-qualitative} shows a test sample from the MultiviewX dataset and the associated label and predictions. In addition to the aforementioned improvements, MVUDA also successfully detects pedestrians that the baseline fails to identify.

\subsection{Baseline data augmentation}
In the main paper, we analyzed the effectiveness of the data augmentations Dropview (DV) \cite{gmvd}, 3D random occlusion (3DR) \cite{3drom}, and the two-level data augmentation (MVA) by \cite{mvaug} in the context of self-training. In this section, we instead analyze their impact on the generalization capability of the baseline, which is trained as described in the main paper. In \cref{tab:baseline}, it can be seen that both DV and 3DR improves the performance on six out of the seven benchmarks. However, MVA improves performance only on four benchmarks. Therefore, we evaluate all three augmentation methods together and ablate MVA in the rightmost columns. Evidently, DV and 3DR yields the overall best performance, while adding MVA typically degrades performance. The findings here are similar to those of the main paper, indicating that the simple data augmentation methods DV and 3DR outperform MVA in the domain generalization setting.

\begin{table}[]
\centering
\begin{tabular}{llllll}
\hline
w/o       & DV        & MVA      & 3DR      & All      & DV+3DR    \\ \hline
\multicolumn{6}{c}{MultiviewX $\rightarrow$ WildTrack}      \\
72.4      & \textbf{73.2}      & 67.1     & 70.4     & 67.8     & 70.0      \\ \hline
\multicolumn{6}{c}{WildTrack $\rightarrow$ MultiviewX}      \\
32.0      & 35.0      & 30.1     & \textbf{36.1}     & 32.1     & 35.9      \\ \hline
\multicolumn{6}{c}{Wildtrack 2,4,5,6 $\rightarrow$ 1,3,5,7} \\
68.7      & 70.0      & 71.3     & 74.6     & 72.3     & \textbf{75.2}      \\ \hline
\multicolumn{6}{c}{Wildtrack 1,3,5,7 $\rightarrow$ 2,4,5,6} \\
62.1      & 65.5      & 59.6     & 66.2     & 66.8     & \textbf{72.3}      \\ \hline
\multicolumn{6}{c}{MultiviewX 1,2,6$\rightarrow${}3,4,5}    \\
46.2      & 51.2      & 52.5     & 52.5     & 53.7     & \textbf{54.7} \\ \hline
\multicolumn{6}{c}{GMVD1 $\rightarrow$ MultiviewX}          \\
60.5      & 65.1      & 64.3     & \textbf{70.8}     & 70.7     & 70.3      \\ \hline
\multicolumn{6}{c}{GMVD2 $\rightarrow$ MultiviewX}          \\
60.0      & 57.6      & 65.4     & 64.7     & \textbf{68.4}     & 66.9       
\end{tabular}
\caption{Performance (MODA) of the baseline trained with different data augmentation methods.}
\label{tab:baseline}
\end{table}

\subsection{Hyperparameter $\lambda$}
In this section, we study how the weight $\lambda$ effects self-training. In \cref{tab:lambda}, the performance of self-training is shown on two benchmarks. Here, self-training is done with local-max pseudo-labeling with $k_d=3$, $\alpha=0.99$ and no data augmentation. It can be seen that the performance is relatively insensitive to the choice of $\lambda$. Conveniently, $\lambda=1$ works well on both benchmarks and yields the best overall performance (tied with $\lambda=2.0$).

\begin{table}[]
\centering
\begin{tabular}{lllll}
\hline
$\lambda=$      & 0.1       & 0.5       & 1.0      & 2.0      \\ \hline
\multicolumn{5}{c}{MultiviewX $\rightarrow$ Wildtrack (70.0)} \\ \hline
                & 75.4      & 77.7      & \textbf{79.7}     & 79.1     \\ \hline
\multicolumn{5}{c}{GMVD1 $\rightarrow${}MultiviewX (70.3)}    \\ \hline
                & 85.7      & 87.1      & 87.8     & \textbf{88.4}    
\end{tabular}
\caption{Performance (MODA) of self-training on two benchmarks with different values of $\lambda$.}
\label{tab:lambda}
\end{table}

\subsection{Perspective supervision}
\begin{figure}
    \centering
    \includegraphics[width=1.0\linewidth]{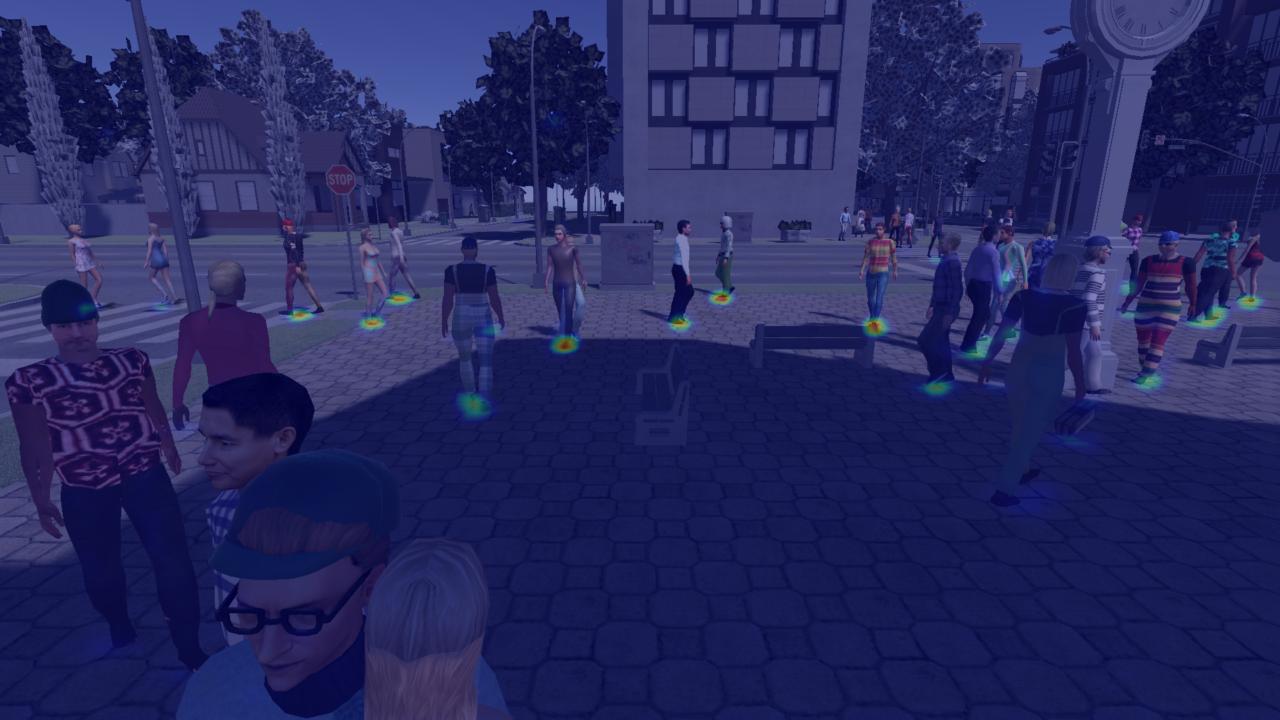}
    \caption{Example of the regressed foot heat map $\hat{y}^n_{f}$ for the first camera ($n=1$) in the MultiviewX dataset.}
    \label{fig:foot_regress}
\end{figure}

\begin{figure}
    \centering
    \includegraphics[width=1.0\linewidth]{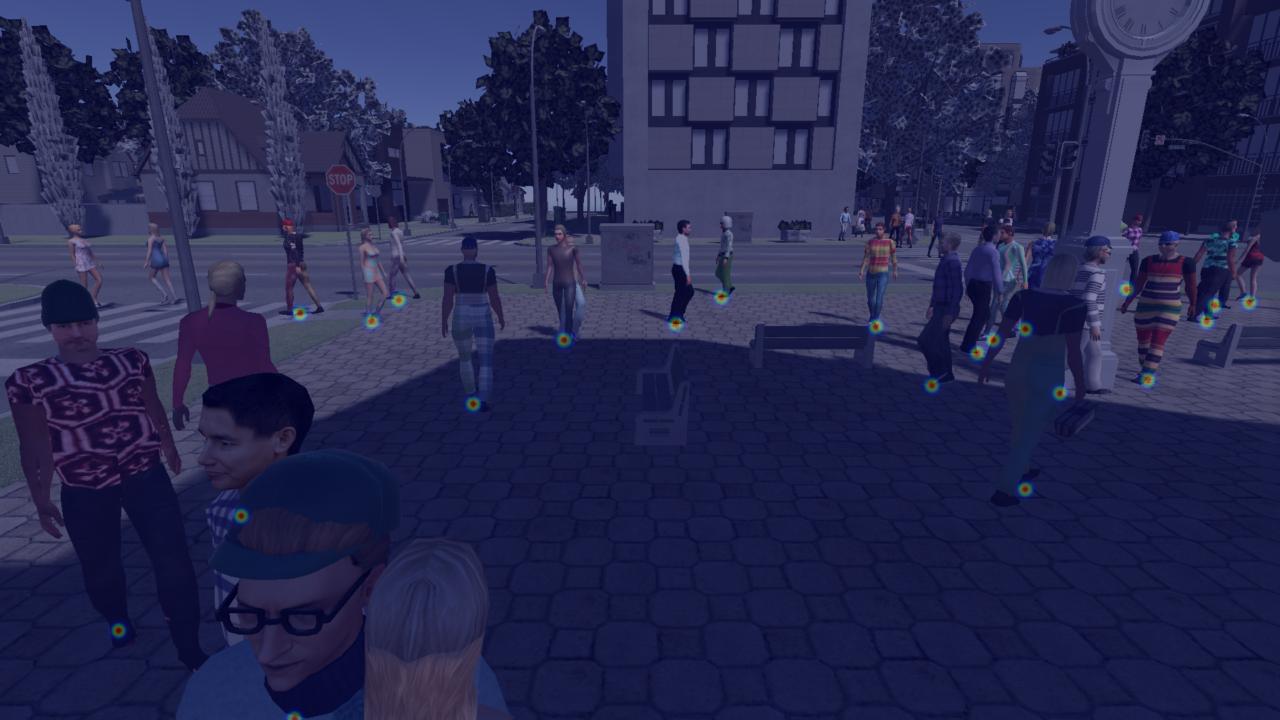}
    \caption{Example of projected pseudo-label $y^n_{f}$ for the first camera ($n=1$) in the MultiviewX dataset.}
    \label{fig:foot_pseudo}
\end{figure}

Following MVDet \cite{mvdet}, we also experimented with training the model on the auxiliary task of single view head-foot detection. To this end, an auxiliary classifier consisting of two convolutional layers is deployed to regress the head and foot positions in each view. Given the single-view features of the n:th view, produced by ResNet-18, it regresses two heat maps $\hat{y}^n_{h}$ and $\hat{y}^n_{f}$ for the head and foot positions of all pedestrians. For illustration, an example of $\hat{y}^n_{f}$ is shown in \cref{fig:foot_regress}. To train this classifier, the positions of the pedestrians in the BEV occupancy map, given by either a label or pseudo-label, are projected into each camera view to create perspective view labels $y^n_{h}$ and $y^n_{f}$ for the head and foot positions respectively. \Cref{fig:foot_pseudo} illustrates the projected pseudo-label $y^n_{f}$ for a sample in MultiviewX. The projection is pre-computed for all positions on the occupancy grid in the used datasets \cite{chavdarova2018wildtrack,mvdet,gmvd}, and a fixed height of pedestrians is used to get the head position. We refer to the respective datasets for further details. Given the acquired foot and head labels $y^n_{h}$ and $y^n_{f}$ of each of the $N$ views, the perspective view loss $\mathcal{L}^{p}$ is computed as
\begin{equation}
        \begin{aligned}
            \frac{1}{N} \sum_{n=1}^N
         \mathcal{L}_{\mbox{MSE}}(y^n_{h}, \hat{y}^n_{h})+ 
     \mathcal{L}_{\mbox{MSE}}(y^n_{f}, \hat{y}^n_{f}).
    \end{aligned}
\end{equation}
Here, $\mathcal{L}_{\mbox{MSE}}$ denotes the MSE-loss with a Gaussian kernel $G$ as in the main paper, according to
    \begin{equation}
     \mathcal{L}_{\mbox{MSE}}(y, \hat{y}) = 
     \sum_{i=1}^{H} \sum_{j=1}^{W} (G(y^{}_{ij}) - \hat{y}^{}_{ij})^2.
    \label{eq:mse_loss_supp}
    \end{equation}


Following \cite{mvdet}, the total loss is computed by adding the perspective view loss to the BEV loss. In our case, we implement perspective view supervision both on source and target domain using labels and pseudo-labels respectively and therefore add $\mathcal{L}^p$ to $\mathcal{L}^S$ and $\mathcal{L}^T$ defined in \cref{mse_loss}.

In \cref{tab:baselin-persp}, we analyze how training with perspective supervision affects the baseline model's generalization capability. The baseline is trained with the same configuration as in the main paper, although, no data augmentation is used. It can be seen that adding perspective supervision leaves the performance of the baseline relatively unaffected, except for Wildtrack 1,3,5,7$\rightarrow$2,4,5,6 where performance degraded significantly, and GMVD1$\rightarrow$MultiviewX where the performance was greatly increased. Mostly, however, perspective supervision yields very modest improvements. Given the overhead in computation and complexity, it may not be worthwhile to use it in a domain generalization setting. This may explain why GMVD \cite{gmvd} opted not to use it. 
After these experiments, we investigated whether perspective view supervision is more beneficial on the target domain in self-training. \Cref{tab:uda-persp} shows the performance of self-training with or without perspective view supervision, applied only to the target domain. Here, self-training is implemented with local-max pseudo-labeling with $k_d=3$, $\alpha=0.99$, $\lambda=1$, and no data augmentation.
In this case, perspective view supervision results in a slight decrease in MODA on most benchmarks, indicating that perspective supervision on the target domain is not beneficial.

\begin{table}[h!]
\centering
\tabcolsep=4.5pt
\begin{tabular}{lllll}
\hline
\multicolumn{1}{l|}{Method}         & MODA & MODP & Precision & Recall \\ \hline
\multicolumn{5}{c}{MultiviewX $\rightarrow$ Wildtrack}                 \\ \hline
\multicolumn{1}{l|}{Baseline}       & \textbf{72.4} & 73.4 & \textbf{92.6}      & 78.7   \\
\multicolumn{1}{l|}{Baseline+persp} & 72.2 & \textbf{74.1} & 90.9      & \textbf{80.1}   \\ \hline
\multicolumn{5}{c}{Wildtrack $\rightarrow${}MultiviewX}                \\ \hline
\multicolumn{1}{l|}{Baseline}       & 32.0 & 68.7 & \textbf{87.2}      & 37.5   \\
\multicolumn{1}{l|}{Baseline+persp} & \textbf{33.3} & \textbf{69.5} & 86.5      & \textbf{39.5}  \\ \hline
\multicolumn{5}{c}{Wildtrack 2,4,5,6 $\rightarrow$ 1,3,5,7}            \\ \hline
\multicolumn{1}{l|}{Baseline}       & 68.7 & \textbf{72.1} & 89.5      & \textbf{77.8}   \\
\multicolumn{1}{l|}{Baseline+persp} & \textbf{68.9} & 69.9 & \textbf{94.3}      & 73.3   \\ \hline
\multicolumn{5}{c}{Wildtrack 1,3,5,7 $\rightarrow$ 2,4,5,6}            \\ \hline
\multicolumn{1}{l|}{Baseline}       & \textbf{62.1} & \textbf{67.5} & \textbf{89.6}      & 70.3   \\
\multicolumn{1}{l|}{Baseline+persp} & 56.9 & 66.0 & 83.0      & \textbf{71.5}   \\ \hline
\multicolumn{5}{c}{MultiviewX 1,2,6$\rightarrow${}3,4,5}               \\ \hline
\multicolumn{1}{l|}{Baseline}       & 46.2 & 68.4 & 82.5      & \textbf{58.6}   \\
\multicolumn{1}{l|}{Baseline+persp} & \textbf{47.7} & \textbf{69.4} & \textbf{85.7}      & 57.3  \\ \hline
\multicolumn{5}{c}{GMVD1 $\rightarrow$ MultiviewX}              \\ \hline
\multicolumn{1}{l|}{Baseline}       & 60.5 & 73.5 & \textbf{90.3}      & 67.8   \\
\multicolumn{1}{l|}{Baseline+persp} & \textbf{66.0} & \textbf{74.1} & 90.2      & \textbf{74.0}   \\ \hline
\multicolumn{5}{c}{GMVD2 $\rightarrow$ MultiviewX}              \\ \hline
\multicolumn{1}{l|}{Baseline}       & 60.0 & \textbf{73.2} & \textbf{91.3}      & 66.3   \\
\multicolumn{1}{l|}{Baseline+persp} & \textbf{60.3} & 72.2 & 86.7      & \textbf{71.3}   
\end{tabular}
\caption{Performance comparison of the baseline trained with or without perspective view supervision.}
\label{tab:baselin-persp}
\end{table}

\begin{table}[h!]
\centering
\tabcolsep=4.5pt
\begin{tabular}{lllll}
\hline
\multicolumn{1}{l|}{Method}    & MODA & MODP & Precision & Recall \\ \hline
\multicolumn{5}{c}{MultiviewX $\rightarrow$ Wildtrack}            \\ \hline
\multicolumn{1}{l|}{UDA}  & \textbf{76.8} & \textbf{74.8} & 92.5      & \textbf{83.5}   \\
\multicolumn{1}{l|}{UDA persp} & 75.5 & 74.5 & \textbf{93.2}      & 81.5   \\ \hline
\multicolumn{5}{c}{Wildtrack $\rightarrow${}MultiviewX}           \\ \hline
\multicolumn{1}{l|}{UDA}  & \textbf{73.1} & 71.5 & \textbf{89.3}      & 83.0   \\
\multicolumn{1}{l|}{UDA persp} & 72.8 & \textbf{72.3} & 86.7      & \textbf{85.9}  \\ \hline
\multicolumn{5}{c}{Wildtrack 2,4,5,6 $\rightarrow$ 1,3,5,7}       \\ \hline
\multicolumn{1}{l|}{UDA}  & 78.0 & 73.7 & \textbf{96.6}      & 80.9   \\
\multicolumn{1}{l|}{UDA persp} & \textbf{78.2} & \textbf{76.2} & 94.3      &\textbf{83.2}   \\ \hline
\multicolumn{5}{c}{Wildtrack 1,3,5,7 $\rightarrow$ 2,4,5,6}       \\ \hline
\multicolumn{1}{l|}{UDA}  & \textbf{79.9} & 70.9 & \textbf{96.2}      & 83.2   \\
\multicolumn{1}{l|}{UDA persp} & \textbf{79.9} & \textbf{71.3} & 92.7      & \textbf{86.8}   \\ \hline
\multicolumn{5}{c}{MultiviewX 1,2,6$\rightarrow${}3,4,5}          \\ \hline
\multicolumn{1}{l|}{UDA}         & \textbf{62.9} & 72.5 & 90.2      & \textbf{70.5 }  \\
\multicolumn{1}{l|}{UDA persp}        & 62.8 & \textbf{72.8} & \textbf{90.4}      & 70.2 \\ \hline
\multicolumn{5}{c}{GMVD1 $\rightarrow$ MultiviewX}         \\ \hline
\multicolumn{1}{l|}{UDA}         & \textbf{88.0} & \textbf{78.3} & 96.5      & \textbf{91.4}   \\
\multicolumn{1}{l|}{UDA persp}        & 87.3 & 77.5 & \textbf{96.6}      & 90.6   \\ \hline
\multicolumn{5}{c}{GMVD2 $\rightarrow$ MultiviewX}         \\ \hline
\multicolumn{1}{l|}{UDA}         & \textbf{87.9} & 76.8 & 96.7      & \textbf{91.0 }  \\
\multicolumn{1}{l|}{UDA persp}        & 87.8 & \textbf{78.3} & \textbf{97.4}      & 90.2   
 
\end{tabular}
\caption{Performance comparison of self-training with or without perspective view supervision.}
\label{tab:uda-persp}
\end{table}

\subsection{Longer training}
In the main paper, we found that MVUDA may benefit from longer trainings, especially when the mean teacher is evolving slowly (high $\alpha$). Therefore, in this section, we present the results of training MVUDA for 20 epochs with $\alpha=0.999$. The rest of the training configuration is kept the same as when training MVUDA for 5 epochs in \cref{tab:main-res,tab:main-res-uncommon} of the main paper. Importantly, we use data augmentation in these experiments, which was not included when the parameter $\alpha$ was studied in the main paper. In \cref{tab:longer}, we compare the performance of MVUDA, which is trained for 5 epochs with $\alpha=0.99$, and MVUDA (ext), trained for 20 epochs with $\alpha=0.999$. It can be seen that MVUDA (ext) achieves higher MODA on all benchmarks except on MultiviewX$\rightarrow$Wildtrack, showing that substantially longer trainings typically result in higher performance. On Wildtrack 1,3,5,7$\rightarrow$2,3,5,6 and MultiviewX 1,2,6$\rightarrow$3,4,5, the performance gain is substantial, with an increase of $3.5$ and $4.7$ MODA respectively. Meanwhile, the results on the first benchmark demonstrate that a slowly evolving teacher is not always beneficial. Rather, it risks converging to a suboptimal local minimum that could have been avoided had the mean teacher been updated more rapidly.

\begin{table}[]
\centering
\tabcolsep=4.5pt
\begin{tabular}{lllll}
\hline
\multicolumn{1}{l|}{Method}    & MODA & MODP & Precision & Recall \\ \hline
\multicolumn{5}{c}{MultiviewX $\rightarrow$ Wildtrack}         \\ \hline
\multicolumn{1}{l|}{MVUDA}      & \textbf{85.4} & 75.3 & \textbf{96.5}      & 88.7   \\
\multicolumn{1}{l|}{MVUDA (ext)}     & 82.9 & \textbf{76.4} & 91.1      & \textbf{91.8}   \\ \hline
\multicolumn{5}{c}{Wildtrack $\rightarrow${}MultiviewX}        \\ \hline
\multicolumn{1}{l|}{MVUDA}      & 82.4 & \textbf{75.4} & 93.3      & 88.8   \\
\multicolumn{1}{l|}{MVUDA (ext)}     & \textbf{83.6} & 74.8 & \textbf{93.7}      & \textbf{89.6}  \\ \hline
\multicolumn{5}{c}{Wildtrack 2,4,5,6 $\rightarrow$ 1,3,5,7}       \\ \hline
\multicolumn{1}{l|}{MVUDA}  & \textbf{79.4} & \textbf{77.8} & \textbf{96.3}      & 82.6   \\
\multicolumn{1}{l|}{MVUDA (ext)} & \textbf{79.4} & 74.9 & 95.8      & \textbf{83.1}   \\ \hline
\multicolumn{5}{c}{Wildtrack 1,3,5,7 $\rightarrow$ 2,4,5,6}       \\ \hline
\multicolumn{1}{l|}{MVUDA}  & 81.4 & 68.8 & \textbf{95.9}      & 85.1   \\
\multicolumn{1}{l|}{MVUDA (ext)} & \textbf{84.9} & \textbf{70.5} & 93.4      & \textbf{91.3}   \\ \hline
\multicolumn{5}{c}{MultiviewX 1,2,6$\rightarrow${}3,4,5}          \\ \hline
\multicolumn{1}{l|}{MVUDA}         & 64.2 & \textbf{73.0} & 91.3      & 71.0   \\
\multicolumn{1}{l|}{MVUDA (ext)}        & \textbf{68.9} & 72.7 & \textbf{92.2}      & \textbf{75.3}  \\ \hline
\multicolumn{5}{c}{GMVD1 $\rightarrow$ MultiviewX}         \\ \hline
\multicolumn{1}{l|}{MVUDA}         & 89.0 & 78.4 & 97.0      & 91.8   \\
\multicolumn{1}{l|}{MVUDA (ext)}        & \textbf{89.8} & \textbf{78.7} & \textbf{97.3}      & \textbf{92.4}   \\ \hline
\multicolumn{5}{c}{GMVD2 $\rightarrow$ MultiviewX}         \\ \hline
\multicolumn{1}{l|}{MVUDA}         & 88.8 & 76.9 & \textbf{97.2}      & 91.5   \\
\multicolumn{1}{l|}{MVUDA (ext)}        & \textbf{90.2} & \textbf{78.7} & 96.7      & \textbf{93.4} 
\end{tabular}
\caption{Performance comparison of MVUDA, which has been trained for 5 epochs with $\alpha=0.99$, and MVUDA (ext), trained for 20 epochs with $\alpha=0.999$.}
\label{tab:longer}
\end{table}

\end{document}